\documentclass[10pt,twocolumn,letterpaper,pagebackref,breaklinks,colorlinks,allcolors=wacvblue]{article}

\usepackage[pagenumbers]{wacv} 
\usepackage{graphicx}
\usepackage{booktabs}
\usepackage{placeins}
\usepackage[accsupp]{axessibility}

\usepackage{tikz}
\usetikzlibrary{calc,shapes.geometric}
\usepackage{colortbl}
\usepackage{arydshln}
\usepackage{multirow}
\usepackage{wrapfig}

\usepackage{orcidlink}
\usepackage{marvosym}

\newcommand{\Paragraph}[1]{\vspace{1mm} \noindent \textbf{#1} \hspace{0mm}}

\definecolor{wacvblue}{rgb}{0.21,0.49,0.74}

\def\wacvPaperID{1001} 
\def\confName{WACV}
\def\confYear{2027}

\title{IDSelect: An RL-based Cost-Aware Selection Agent for Video-based Person Recognition}

\author{
Yuyang Ji$^{1}$\quad
Yixuan Shen$^{1}$\quad
Kien Nguyen$^{3}$\quad
Lifeng Zhou$^{2}$\quad
Feng Liu$^{1}$\\
$^1$ Department of Computer Science, Drexel University\\
$^2$ Department of Electrical and Computer Engineering, Drexel University\\
$^3$ School of Electrical Engineering and Robotics, Queensland University of Technology\\
{\tt\small fl397@drexel.edu}
}

\begin{document}
\maketitle


\begin{abstract}
Video-based person recognition achieves robust identification by integrating face, body, and gait. However, current systems waste computational resources by processing all modalities with fixed heavyweight ensembles regardless of input complexity. To address these limitations, we propose IDSelect, a reinforcement learning-based cost-aware selector that chooses one pre-trained model per modality per-sequence to optimize the accuracy-efficiency trade-off. Our key insight is that input-conditioned selection discovers complementary model choices that surpass fixed ensembles with fewer resources. IDSelect trains a lightweight agent end-to-end using actor-critic reinforcement learning with budget-aware optimization. The reward balances recognition accuracy with computational cost, while entropy regularization prevents premature convergence. At inference, the policy selects the most probable model per modality and fuses modality-specific similarities for the final score. Extensive experiments on challenging video-based datasets demonstrate IDSelect's superior efficiency: on CCVID, it achieves 95.9\% Rank-1 accuracy with 92.4\% less computation than strong baselines while improving accuracy by 1.8\%; on MEVID, it reduces computation by 41.3\% while maintaining comparable Rank-1 accuracy.
\href{https://vilab-group.com/project/idselect}{Project}
\end{abstract}
    
\section{Introduction}
\label{sec:intro}

\begin{figure}[t]
  \centering
   \includegraphics[width=1\linewidth]{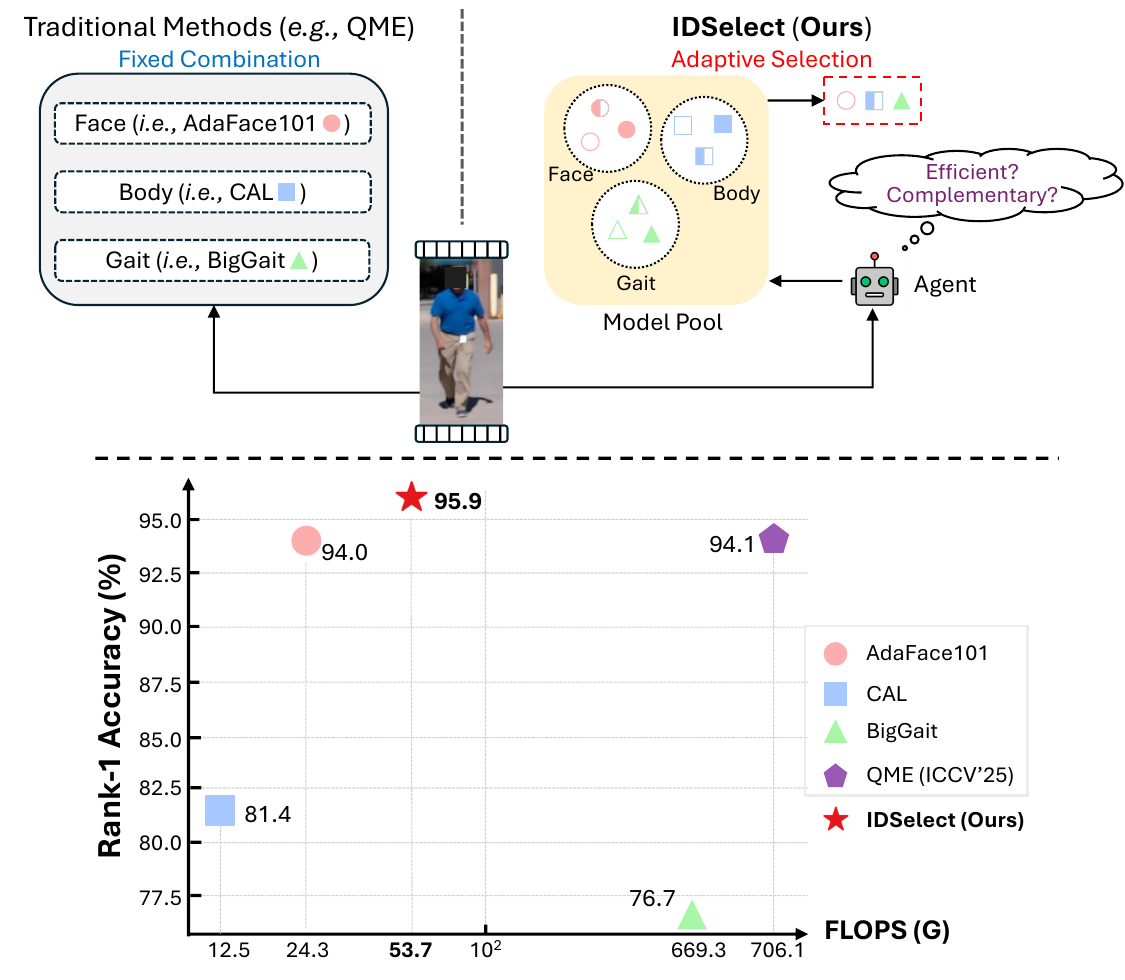}
   \vspace{-7mm}
   \caption{\textbf{Top:} Prior multi-modal person recognition methods (\emph{e.g.}, QME~\cite{zhu2025quality}) use fixed model combinations for all inputs, while our IDSelect employs an RL-based cost-aware agent to adaptively select complementary models from diverse pools based on input characteristics. \textbf{Bottom:} Accuracy vs. computational cost on CCVID dataset~\cite{CAL}. Our method achieves superior accuracy (95.9\%, $+1.8\%$) with 92.4\% fewer FLOPs than QME.}
   \label{fig:teaser}
   \vspace{-3mm}
\end{figure}

Whole-body person recognition~\cite{liu2024farsight,jawade2024conan,huang2024whole,liu2025person}, integrating facial appearance, gait patterns, and body shape, has emerged as a robust approach for identifying individuals in unconstrained video environments.
By fusing complementary biometric cues, these systems achieve superior accuracy in scenarios where single-modality methods fail (\emph{e.g.}, low-light or occluded conditions).
Prior whole-body systems~\cite{huang2024whole,liu2025person} consistently demonstrate that these three cues provide the strongest complementarity: faces are highly discriminative when resolution and pose are favorable, full-body appearance captures clothing and coarse shape, and gait remains reliable when faces are small, blurred, or occluded.
This capability is crucial for applications in law enforcement, border security, surveillance, and public media analytics~\cite{gong2011person,zheng2016person,jain2024speaker,nguyen2026aerialsurvey,nguyen2025agvpreid}.

%

However, current state-of-the-art video-based multi-modal systems~\cite{liu2025person} are computationally impractical for real-world deployment: they typically use a \emph{\textbf{fixed combination}} of heavy models for all modalities, incurring high memory and latency costs regardless of input difficulty. In resource-constrained settings like edge devices, processing every modality with large networks becomes prohibitively slow and power-intensive. This \emph{\textbf{one-size-fits-all}} strategy ignores deployment budgets and fails to exploit the fact that not all inputs require powerful (and expensive) models.

Recent research has begun to expose these inefficiencies. The quality-guided fusion method~\cite{zhu2025quality}, for instance, dynamically reweights modality contributions based on biometric sample quality, yielding accuracy gains under challenging conditions.
However, such methods still assume a fixed pool of models and overlook the computational burden of always running multiple high-capacity networks. 
While some works~\cite{li2019airface,george2024edgeface} have explored using lighter vs. heavier models within a single modality, such as creating lightweight face recognition architectures that reduce computational requirements, these approaches remain limited to single modality optimization and do not address the more complex challenge of joint optimization across modalities, \emph{i.e.}, discovering which combination of face, gait, and body models achieves optimal accuracy-cost trade-offs. 
 \emph{A unified agent that dynamically adapts video-based multi-modal inference to varying computational budgets while discovering effective cross-modal synergies is highly desired.}

To address these limitations, we introduce \textbf{IDSelect}, an RL-based cost-aware agent that dynamically selects one pre-trained model per modality for each input. Our key insight is that a learned selection agent can discover complementary, input-conditioned model combinations that outperform static ensembles while utilizing significantly fewer computational resources. 
Unlike prior RL-based adaptive inference methods~\cite{bolukbasi2017adaptive,rao2017attention} that target single-modality early exit within one network, IDSelect addresses the fundamentally different problem of cross-modal expert selection for retrieval. This introduces unique technical challenges absent in prior work: (1) \emph{probe-gallery consistency}— the same model must be applied to both probe and gallery sequences per modality for valid similarity computation, since different models produce incompatible embeddings; (2) the \emph{combinatorial cross-modal action space} spans heterogeneous face, gait, and body pools with distinct failure modes, where we show that cross-modal diversity matters more than intra-modal redundancy.
Importantly, we choose to select only one model per modality rather than multiple models within the same modality. This design choice is motivated by two key observations: First, models within the same modality tend to learn similar feature representations~\cite{jain2005score,nandakumar2008likelihood}, leading to high redundancy and diminishing marginal returns when combined. Second, true complementarity arises from \emph{cross-modal} diversity; face, gait, and body capture fundamentally different biometric signatures—whereas selecting multiple face (or gait, or body) models primarily increases computational cost without proportional accuracy gains. 
By intelligently selecting model combinations based on input characteristics and budget constraints, IDSelect achieves competitive recognition performance at a significantly reduced computational cost.

Specifically, we first curate diverse pools of pre-trained models for each modality, spanning multiple architectures and training corpora to expose distinct cost–accuracy trade-offs. 
IDSelect employs a lightweight selection agent over a discrete video-based multi-modal action space, selecting \emph{one model per modality} for each sample. 
During training, we construct positive/negative pairs and optimize an end-to-end hybrid objective that balances recognition accuracy, computational efficiency, and policy diversity. 
To make discrete selections learnable, we train the policy via an actor-critic reinforcement learning framework with a Lagrangian budget controller, which enforces a target computation limit while balancing reward and cost. We also employ a budget curriculum that gradually tightens computational constraints to improve robustness across operating points. 
At inference, IDSelect adaptively determines optimal model combinations and aggregates per-modality similarities for final recognition. This mechanism enables IDSelect to discover effective model synergies while adapting computation to input characteristics.

In summary, the contributions of this work include:

$\diamond$ We propose \textbf{IDSelect}, an RL-based cost-aware agent that discovers complementary model combinations across face, gait, and body modalities for efficient whole-body recognition.

$\diamond$ We devise a lightweight selection agent trained end-to-end via an actor-critic reinforcement learning framework with a Lagrangian budget controller, enabling stable and budget-aware model selection.

$\diamond$ Extensive experiments on video-based whole-body datasets demonstrate that IDSelect discovers superior model combinations, achieving better accuracy-cost trade-offs than fixed and quality-guided fusion baselines.
\section{Related work}
\label{sec:related_work}

\Paragraph{Whole-Body Person Recognition.}
%
Video-based multi-modal person recognition integrates complementary biometric traits (face, gait, and body) for robust identification in challenging environments~\cite{liu2024farsight,jawade2024conan, huang2024whole,nguyen2026aerialsurvey}. Unlike uni-modal approaches relying on single biometric sources~\cite{deng2019arcface,kim2022adaface, fan2023opengait, kim2024keypoint,CAL,kim2025sapiensid,su2026sapiensid2,kim2022cluster,liu2022cfsm}, whole-body systems mitigate modality weaknesses through cross-modal complementarity. Face recognition struggles with extreme pose variations~\cite{liu2023learning,liu2024distilling,su2025hamobe,nguyen2025agvpreid}; gait analysis is sensitive to clothing changes, and body shape can be affected by posture, yet their combination provides robustness where individual modalities fail~\cite{zhang2020learning}. Recent advances have explored end-to-end pipelines integrating detection, restoration, and recognition~\cite{liu2024farsight,liu2025person}, demonstrating improved performance in unconstrained settings. However, these systems deploy all models simultaneously, leading to substantial computational overhead regardless of input complexity.
%

\Paragraph{Video-based Multi-Modal Biometric Fusion.}
Traditional video-based multi-modal fusion relies on score-level strategies, where similarity scores from individual modalities are normalized and aggregated using simple operations~\cite{jain2005score}. More advanced methods adopt likelihood ratio-based fusion~\cite{nandakumar2008likelihood} for probabilistic interpretability or feature-level fusion~\cite{aung2022multimodal} to capture cross-modal correlations. Recent works further explore complementary and adaptive multi-modal representations: SHaRC~\cite{zhu2024sharc} combines shape and appearance cues for person identification, while FlexiReID~\cite{sun2025flexireid}
uses adaptive mixture-of-experts for multi-modal representation integration. Zhu \emph{et al.}~\cite{zhu2025quality} proposed the Quality-guided Mixture of score-fusion Experts (QME) framework, introducing pseudo-quality loss for quality estimation and score triplet loss for improved metric learning. The framework dynamically weights modality contributions based on quality estimates, addressing cost-effectiveness by reducing overhead on poor-quality inputs. While the approach yields superior accuracy through quality-awareness, it fundamentally assumes \emph{fixed} model pools and overlooks the burden of executing multiple heavyweight networks simultaneously.

\begin{figure*}[t]
  \centering
   \includegraphics[width=0.9\linewidth]{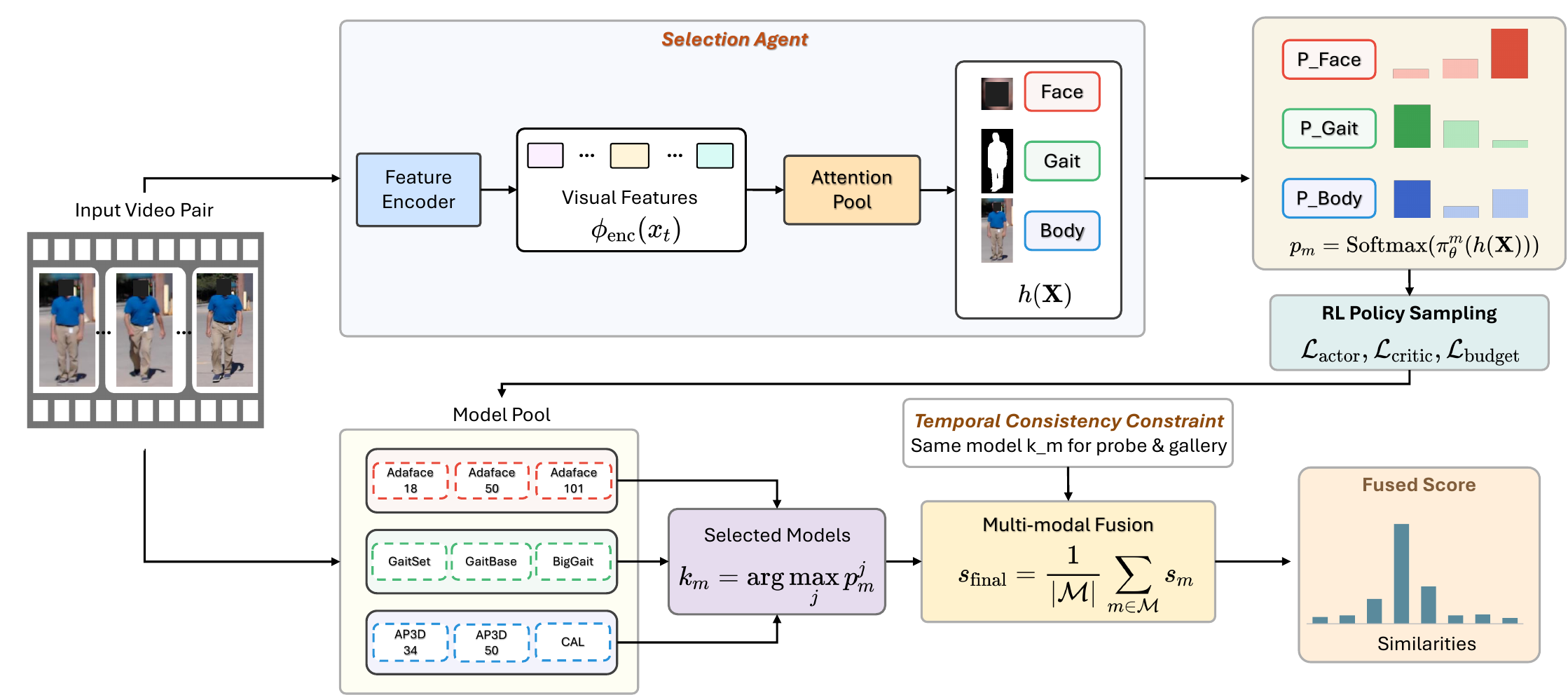}
   \vspace{-2mm}
\caption{IDSelect framework architecture. The selection agent processes input video pairs through a feature encoder and attention pooling to generate modality-specific selection distributions. An actor-critic reinforcement learning policy optimizes model selection from pre-trained pools under a budget constraint. The framework minimizes a multi-objective loss combining classification reward, computational cost, and selection diversity to discover optimal model combinations for multi-modal fusion.}
   \label{fig:flowchart}
   \vspace{0mm}
\end{figure*}

\Paragraph{Adaptive and Cost-Aware Inference.}
The computer vision community has developed sophisticated methods for input-dependent computation. Bolukbasi \emph{et al.}~\cite{bolukbasi2017adaptive} proposed adaptive neural networks that learn policies to route inputs through networks of varying capacity, while Rao \emph{et al.}~\cite{rao2017attention} applied deep RL to attend to informative frames for video face recognition. Early-exit networks dynamically terminate inference when confidence thresholds are met. Teerapittayanon \emph{et al}.~\cite{teerapittayanon2016branchynet} established the foundational BranchyNet framework for confidence-based early exits, while Yang \emph{et al}.~\cite{yang2020resolution} demonstrated spatial redundancy exploitation via Resolution Adaptive Networks. Recent work on Adaptive Vision Transformers has pushed boundaries further: Yin \emph{et al}.~\cite{yin2022vit} developed A-ViT, achieving significant throughput improvements with minimal accuracy drop through adaptive token reduction, and Meng \emph{et al}.~\cite{meng2022adavit} introduced fine-grained control over computational resources via learned usage policies. In video-based multi-modal settings, Panda \emph{et al}.~\cite{panda2021adamml} proposed AdaMML for video understanding, achieving computational reduction through policies that adaptively select modality subsets. However, these adaptive approaches have not been systematically applied to multi-modal biometric recognition.


\Paragraph{Dynamic Model Selection.}
%
Recent developments in efficient model design provide implementations of cost-aware architectures. George \emph{et al}.~\cite{george2024edgeface} introduced EdgeFace, combining a CNN-Transformer hybrid architecture with Low Rank Linear modules, achieving exceptional results with minimal parameters. Yan \emph{et al}.~\cite{yan2019vargfacenet} demonstrated variable group convolutions for adaptive computational complexity in VarGFaceNet. In video-based multi-modal learning, He \emph{et al}.~\cite{he2024efficient} developed theoretical frameworks for quantifying modality utility using greedy submodular maximization, while recent work on Flex-MoE~\cite{yun2024flex} addresses arbitrary modality combination with missing data robustness. However, existing methods either operate within single modalities or assume fixed model combinations, failing to discover complementary model synergies that could simultaneously improve accuracy and reduce computational cost. Reinforcement learning has also been explored for adaptive tool routing and dynamic inference~\cite{zhang2025router}, where an agent learns to select tools or sub-models based on inputs. Inspired by this line of work, our IDSelect agent addresses this gap by learning to discover optimal model combinations across face, gait, and body modalities through an actor-critic reinforcement learning policy that adaptively balances accuracy and computational cost.

\section{Method}
\label{sec:method}

\subsection{Preliminaries}
Multi-modal person recognition leverages temporal sequences to extract complementary biometric traits for robust identification across diverse scenarios. 
Formally, let $\mathcal{I}$ denote the set of all identities, and consider an input video sequence $\mathbf{X} = \{x_1, x_2, \ldots, x_T\}$ containing visual information from multiple biometric modalities $\mathcal{M} = \{\text{face}, \text{gait}, \text{body}\}$ over $T$ frames. Each modality provides a distinct temporal view: facial appearance captures identity-specific features across frames, gait patterns encode behavioral dynamics through walking motion, and body shape provides geometric cues that remain stable throughout the sequence.
The video-based recognition pipeline operates in two phases: \emph{enrollment} and \emph{recognition}. 
During enrollment, gallery video sequences $\mathcal{G} = \{\mathbf{X}_1^g, \mathbf{X}_2^g, \ldots, \mathbf{X}_{|\mathcal{I}|}^g\}$ representing known identities are processed to extract representative temporal features. During recognition, a probe/query video $\mathbf{X}^p$ is compared against the gallery through sequence-level similarity matching.

For each modality $m \in \mathcal{M}$, a temporal feature extractor $f_m: \mathbb{R}^{T \times d_m} \rightarrow \mathbb{R}^{e_m}$ maps the modality-specific video sequence $\mathbf{X}_m = \{x_{1,m}, x_{2,m}, \ldots, x_{T,m}\}$ to a unified embedding space, where $d_m$ represents the frame-level input dimensionality and $e_m$ represents the final embedding dimension. The recognition process computes similarities between probe and gallery sequence embeddings:
\begin{equation}
s_m(\mathbf{X}^p, \mathbf{X}^g) = \frac{f_m(\mathbf{X}_m^p) \cdot f_m(\mathbf{X}_m^g)}{\|f_m(\mathbf{X}_m^p)\| \|f_m(\mathbf{X}_m^g)\|}.
\end{equation}
Multi-modal fusion combines temporal evidence from all modalities to produce the final recognition score. Traditional approaches employ fixed strategies: score-level fusion through weighted averaging or feature-level fusion via joint embeddings; however, they struggle with varying input quality across modalities. Recent quality-guided methods like QME~\cite{zhu2025quality} address this through mixture-of-experts architectures that dynamically weight modalities based on estimated reliability, achieving superior performance by emphasizing high-quality inputs while downweighting degraded ones. However, these approaches assume \emph{fixed} model availability and \emph{presuppose that complementary information is accessible through pre-selected model combinations}, without addressing the computational overhead of executing all candidate networks, regardless of input complexity or deployment constraints.

\subsection{IDSelect Architecture}
We reformulate video-based person recognition as a \emph{dynamic} resource allocation problem, where model combinations are discovered through learnable policies. The framework transforms the recognition pipeline from a static graph into an adaptive inference that balances recognition performance with computational efficiency.
The core innovation lies in decoupling model availability from model utilization. IDSelect introduces the hypothesis that different video sequences exhibit varying complexity patterns that can be effectively handled by different model combinations. This leads to a combinatorial optimization formulation where the selection agent $\pi$ must solve:
\vspace{-2mm}
\begin{equation}
\begin{aligned}
\pi^*(\mathbf{X}) 
= \arg\max_{\pi} \Bigg[ 
   & \sum_{m \in \mathcal{M}} 
     \mathbb{E}[s_m(\mathbf{X}^p, \mathbf{X}^g \mid f_m^{\pi(m)})] \\
   & - \mu \cdot \text{Cost}(\pi(\mathbf{X})) \Bigg],
\end{aligned}
\end{equation}
where $f_m^{\pi(m)}$ represents the selected model for modality $m$, explicitly balancing recognition performance with computational cost to discover model synergies.

As illustrated in Fig.~\ref{fig:flowchart}, IDSelect implements this using three interconnected components. The framework processes videos $\mathbf{X}$ through a selection agent that analyzes cross-modal characteristics, applies reinforcement-driven selection mechanisms to choose model combinations, and ensures temporal coherence in feature extraction.

\Paragraph{Input-Adaptive Selection Agent.}
The selection agent constructs a video-level representation by processing the entire video sequence:
\begin{equation}
h(\mathbf{X}) = \text{AttentionPool}(\{\phi_{\text{enc}}(x_t) : t = 1, 2, \ldots, T\}),
\end{equation}
where $\phi_{\text{enc}}$ is a lightweight CNN encoder that extracts discriminative frame-level features optimized for model selection decisions, and AttentionPool learns to weight frames based on their informativeness for selection. This comprehensive temporal analysis captures sequence-level characteristics, including pose dynamics, motion complexity, visual quality variations, and modality-specific degradation patterns, enabling the agent to predict which model combinations will be most effective for the given input complexity. The resulting representation $h(\mathbf{X})$ encodes both local frame-level properties and global sequence statistics necessary for informed model selection.

Modality-specific policy heads generate selection distributions:
\begin{equation}
p_m = \text{Softmax}(\pi_\theta^m(h(\mathbf{X}))), \quad m \in \{\text{face}, \text{gait}, \text{body}\}.
\end{equation}
The key insight is that optimal model combinations exhibit complementary properties. When facial information is degraded, the policy compensates by selecting more sophisticated gait and body models, enabling the discovery of non-obvious model synergies.


\Paragraph{RL-Guided Model Selection.}
IDSelect addresses the optimization challenge through a stochastic policy that is directly trained via reinforcement learning. By default, IDSelect selects one model per modality; the MEVID \(1F+2B\) setting is adopted to match the evaluation protocol of prior baselines. During training in this setting, the face policy samples a single model index $a_{\text{face}} \sim \text{Categorical}(p_{\text{face}})$, while the body policy draws two actions without replacement by sequentially sampling from masked categorical distributions. The resulting action set $\mathbf{a} = (a_{\text{face}}, a_{\text{body}}^{(1)}, a_{\text{body}}^{(2)})$ specifies one face model and two body models that are applied consistently to both probe and gallery clips. At inference time, selections become deterministic by choosing the argmax for the face branch and the top-2 logits for the body branch.

\Paragraph{Complementary Feature Extraction.}
IDSelect enforces temporal consistency through a constraint-preserving extraction strategy that addresses a critical challenge in adaptive model selection: ensuring that feature comparisons remain valid across different model choices. The framework requires that the same selected model be applied to both probe and gallery sequences to maintain embedding compatibility:

\begin{equation}
\tilde{\mathbf{z}}_m^p = f_m^{k_m}(\mathbf{X}_m^p), \quad \tilde{\mathbf{z}}_m^g = f_m^{k_m}(\mathbf{X}_m^g),
\end{equation}
where $k_m$ is the index chosen by the policy for modality $m$, at inference we use $k_m=\arg\max_j p_m^j$.

This consistency constraint is essential because different models $f_m^i$ and $f_m^j$ typically produce embeddings in different feature spaces, making direct comparison meaningless. By ensuring that probe and gallery features are extracted using identical models, IDSelect maintains the validity of the cosine similarity computation:

\begin{equation}
s_m = \cos(\tilde{\mathbf{z}}_m^p, \tilde{\mathbf{z}}_m^g) = \frac{\tilde{\mathbf{z}}_m^p \cdot \tilde{\mathbf{z}}_m^g}{\|\tilde{\mathbf{z}}_m^p\| \|\tilde{\mathbf{z}}_m^g\|}.
\end{equation}
Furthermore, within each video sequence, the selected model processes all frames consistently to preserve temporal coherence. This design choice ensures that the temporal modeling capabilities of gait recognition models remain intact while enabling adaptive selection at the sequence level.

The final recognition score aggregates evidence across all modalities through simple averaging:
$
s_{\text{final}} = \frac{1}{|\mathcal{M}|} \sum_{m \in \mathcal{M}} s_m.
$
This aggregation strategy, while simple, proves effective in practice as the selection mechanism automatically balances contribution of modalities by choosing appropriate model complexities rather than relying on learned fusion weights.

\subsection{Model Training}

IDSelect is trained end-to-end through an actor-critic formulation that optimizes a reward balancing recognition accuracy against normalized computational cost. Let $s_{\text{final}}$ denote the fused similarity for a paired sample with the ground-truth label $y \in \{0,1\}$. The reward for action set $\mathbf{a}$ is defined as
\begin{equation}
r = \underbrace{1 - \text{BCE}(\sigma(s_{\text{final}}), y)}_{\text{utility}} - \lambda \, \underbrace{C(\mathbf{a})}_{\text{cost}},
\end{equation}
where $\sigma(\cdot)$ is the sigmoid function, $\text{BCE}$ denotes binary cross-entropy, and $C(\mathbf{a})$ is the average of modality-specific FLOP costs normalized by their modality maxima. We use BCE as a differentiable surrogate for the non-differentiable ranking metrics (Rank-1, mAP): encouraging positive-pair similarities to exceed negative-pair similarities aligns with the ranking objective. The scalar $\lambda$ is a Lagrange multiplier that self-adjusts to satisfy the target budget $C_{\text{target}}$: after each batch, we update $\lambda \leftarrow \max(0, \lambda + \eta ( \bar{C} - C_{\text{target}}))$, where $\bar{C}$ is the observed batch-average cost and $\eta$ is a small step size.

A value head $V_\psi(h(\mathbf{X}))$ predicts the expected reward to stabilize policy learning. The actor loss combines policy gradients with entropy regularization:
\begin{equation}
\mathcal{L}_{\text{actor}} = - (r - V_\psi) \cdot \log \pi_\theta(\mathbf{a} \mid \mathbf{X}) - \beta \, \mathcal{H}(\pi_\theta),
\end{equation}
where $\mathcal{H}$ denotes the entropy of the joint action distribution, and $\beta$ encourages exploration early in training. The critic learns by minimizing
$
\mathcal{L}_{\text{critic}} = (r - V_\psi)^2,
$
and the training objective is $\mathcal{L}_{\text{actor}} + \alpha \mathcal{L}_{\text{critic}}$ 
with $\alpha$ controlling the critic weight. All biometric backbones remain frozen, 
so gradients only update the selection agent, the value head, and the optimizer 
state for the Lagrange multiplier.




\subsection{Implementation Details}

All pre-trained models remain frozen, focusing optimization on agent learning only. During training, we sample video pairs $(\mathbf{X}^p, \mathbf{X}^g)$ with positive and negative identity labels. The selection agent analyzes each query video $\mathbf{X}^p$ to determine model selections, which are applied to both query and gallery videos for feature extraction.

At inference, all candidate models are loaded once at initialization with no runtime swapping. Gallery embeddings are precomputed for each candidate model and indexed using the probe-selected model, avoiding per-query gallery-side recomputation. Detailed architecture, and settings are provided in the supplementary material.




\definecolor{lightred}{RGB}{255,150,150}
\definecolor{lightgreen}{RGB}{150,220,150}
\definecolor{lightblue}{RGB}{150,180,255}

\newlength{\iconsize}
\setlength{\iconsize}{1.6ex}
\newcommand{\iconlw}{0.35pt}

\newcommand{\levfrac}[1]{%
  \ifcase#1 0\or .25\or .5\or .75\or 1\fi
}

\newcommand{\symcircle}[2]{%
\tikz[baseline=-0.6ex]{
  \def\S{\iconsize}
  \def\F{\levfrac{#1}}
  \draw[line width=\iconlw,draw=#2] (0,0) circle (.5*\S);
  \begin{scope}
    \clip (0,0) circle (.5*\S);
    \fill[#2] (-.5*\S,-.5*\S) rectangle ({-.5*\S + \F*\S}, .5*\S);
  \end{scope}
}}

\newcommand{\symsquare}[2]{%
\tikz[baseline=-0.6ex]{
  \def\S{\iconsize}
  \def\F{\levfrac{#1}}
  \draw[line width=\iconlw,draw=#2] (-.5*\S,-.5*\S) rectangle (.5*\S,.5*\S);
  \begin{scope}
    \clip (-.5*\S,-.5*\S) rectangle (.5*\S,.5*\S);
    \fill[#2] (-.5*\S,-.5*\S) rectangle ({-.5*\S + \F*\S}, .5*\S);
  \end{scope}
}}

\newcommand{\symtriangle}[2]{%
\tikz[baseline=-0.3ex]{
  \def\S{\iconsize}
  \def\F{\levfrac{#1}}
  \node[draw=#2,regular polygon,regular polygon sides=3,
        minimum size={1.2*\S}, 
        inner sep=0pt,line width=\iconlw,
        shape border rotate=0] (t) {};
  \begin{scope}
    \path[clip] (t.corner 1)--(t.corner 2)--(t.corner 3)--cycle;
    \fill[#2] let \p1=(t.center) in
      ($(\p1)+(-.5*\S,-.6*\S)$)
      rectangle ($(\p1)+(-.5*\S+\F*\S,.6*\S)$);
  \end{scope}
}}

\newcommand{\faceA}{\symcircle{0}{lightred}}
\newcommand{\faceB}{\symcircle{1}{lightred}}
\newcommand{\faceC}{\symcircle{2}{lightred}}
\newcommand{\faceD}{\symcircle{3}{lightred}}
\newcommand{\faceE}{\symcircle{4}{lightred}}

\newcommand{\bodyA}{\symsquare{0}{lightblue}}
\newcommand{\bodyB}{\symsquare{1}{lightblue}}
\newcommand{\bodyC}{\symsquare{2}{lightblue}}
\newcommand{\bodyD}{\symsquare{3}{lightblue}}
\newcommand{\bodyE}{\symsquare{4}{lightblue}}

\newcommand{\gaitA}{\symtriangle{0}{lightgreen}}
\newcommand{\gaitB}{\symtriangle{1}{lightgreen}}
\newcommand{\gaitC}{\symtriangle{2}{lightgreen}}
\newcommand{\gaitD}{\symtriangle{3}{lightgreen}}
\newcommand{\gaitE}{\symtriangle{4}{lightgreen}}

\newcommand{\gray}{\cellcolor{gray!25}}
\section{Experiments}\label{sec:experiments}

\subsection{Experimental Setup}

\Paragraph{Datasets.}
Following the setting of QME~\cite{zhu2025quality}, we evaluate IDSelect on two challenging video-based person recognition datasets that exhibit significant real-world variations. 
CCVID dataset~\cite{CAL} contains 2,856 sequences across 226 identities, where each identity has 2 to 5 different suits of clothing. The dataset presents challenging variations, including different clothing across sessions, varying illumination conditions, and temporal gaps between captures. 
MEVID dataset~\cite{davila2023mevid} spans extensive indoor and outdoor environments across nine unique dates in a 73-day window, featuring 158 unique people wearing 598 outfits taken from 8,092 tracklets, with an average length of 590 frames captured from 33 camera views. 
The dataset emphasizes richer information per individual, with 4 outfits per identity and 33 viewpoints across 17 locations, inherited from the MEVA person activities dataset, which is intentionally demographically balanced. 

\Paragraph{Model Pools.}
We strategically design diverse model pools to evaluate IDSelect's adaptability across different computational constraints and model capabilities. For CCVID, we evaluate two complementary configurations:

\vspace{1mm}
\noindent (1) \emph{Configuration 1 (QME Baseline Comparison)}: This configuration directly compares against the QME~\cite{zhu2025quality} (ICCV'25) baseline by including the \textbf{\emph{exact same}} heavyweight models as the most expensive options per modality: AdaFace101~\cite{kim2022adaface} (ResNet101 backbone, face), CAL (ResNet50, body), and BigGait (ViT-B16, gait). We additionally include lighter alternatives across all modalities: AdaFace18/50 (ResNet18/50 backbones) for face recognition, AP3D 34/50~\cite{gu2020appearance} (ResNet34/50 backbones) for body analysis, and GaitSet~\cite{chao2021gaitset}/GaitBase~\cite{fan2023opengait} for gait recognition. This design enables fair comparison with QME while demonstrating IDSelect's ability to achieve superior efficiency through intelligent model selection rather than brute-force heavyweight fusion.

\vspace{1mm}
\noindent (2) \emph{Configuration 2 (SOTA Model Diversity)}: To validate generalization beyond QME's specific model choices, we incorporate diverse state-of-the-art architectures, including TF-CLIP~\cite{yu2024tf} for body recognition and DeepGaitv2 for gait analysis, alongside AdaFace variants and complementary models (CAL~\cite{CAL}, AP3D 50~\cite{gu2020appearance}, GaitSet~\cite{chao2021gaitset}, GaitBase~\cite{fan2023opengait}). This configuration tests IDSelect's adaptability to different SOTA architectures while maintaining reasonable computational constraints.


For MEVID's challenging conditions, following the setting of QME~\cite{zhu2025quality}, we focus on face-body fusion using AdaFace variants (AdaFace18/50/101 with ResNet18/50/101 backbones) and five body models: CAL~\cite{CAL}, AGRL~\cite{wu2020adaptive}, Attn-CL~\cite{pathak2020video}, AP3D 50~\cite{gu2020appearance}, and PiT~\cite{zang2022multidirection}. This configuration excludes gait modality due to severely degraded body image resolution, which prevents reliable gait silhouette extraction.


\begin{table}[t]
\centering
\caption{Performance comparison on CCVID \emph{Configuration 1}. IDSelect achieves superior accuracy-efficiency trade-offs compared to single modality baselines, fixed fusion strategies, and quality-guided methods. `Ada.' denotes input-adaptive selection. Colored icons in ``Comb.'' represent modality and model capacity (\textcolor{lightred}{$\circ$}=face, \textcolor{lightblue}{$\square$}=body, \textcolor{lightgreen!70!black}{$\triangle$}=gait; fill level increases with model size, from light to heavy). Models marked with an asterisk * are not finetuned on the CCVID dataset, while all other models are finetuned. F=Face, B=Body, G=Gait. Heavy/Light denotes the highest/lowest-cost model.}
\label{tab:ccvid1}

\setlength{\tabcolsep}{4pt}
\renewcommand{\arraystretch}{1}
\scriptsize  

\begin{minipage}[t]{0.49\textwidth}
\centering
\resizebox{1\linewidth}{!}{
\begin{tabular}{l c c c c c}
\toprule
\textbf{Method} & \textbf{Arch.} & \textbf{GFLOPs$\downarrow$} & \textbf{Comb.} & \textbf{R1$\uparrow$} & \textbf{mAP$\uparrow$} \\
\midrule

\multicolumn{6}{l}{\gray  \emph{\textbf{Face Models}}} \\ 
\hdashline
AdaFace18*~\cite{kim2022adaface}  & R18  &  5.2  & \faceA & 90.5 & 80.4\\
AdaFace50*~\cite{kim2022adaface}  & R50  & 12.7  & \faceC & 91.7 & 82.9\\
AdaFace101*~\cite{kim2022adaface} & R101 & 24.3  & \faceE & 94.0 & 87.9\\[-2pt]
\addlinespace[2pt]

\multicolumn{6}{l}{\gray  \emph{\textbf{Body Models}}} \\
\hdashline
AP3D 34~\cite{gu2020appearance} & R34 &  7.6  & \bodyA & 75.1 & 73.6 \\
AP3D 50~\cite{gu2020appearance} & R50 &  8.5  & \bodyC & 80.9 & 79.2 \\
CAL~\cite{CAL}                  & R50 & 12.5  & \bodyE & 81.4 & 74.7 \\[-2pt]
\addlinespace[2pt]

\multicolumn{6}{l}{\gray  \emph{\textbf{Gait Models}}} \\
\hdashline
GaitSet~\cite{chao2021gaitset}  & CNN-6     &   6.5  & \gaitA & 79.3 & 73.8 \\
GaitBase~\cite{fan2023opengait} & R9        &  71.0  & \gaitC & 64.0 & 61.4 \\
BigGait*~\cite{ye2024biggait}   & ViT  & 669.3  & \gaitE & 76.7 & 61.0 \\
\bottomrule
\end{tabular}
}
\end{minipage}\hfill
\begin{minipage}[t]{0.49\textwidth}
\centering
\resizebox{1\linewidth}{!}{
\begin{tabular}{l c c c c}
\toprule
\textbf{Method} & \textbf{GFLOPs$\downarrow$} & \textbf{Comb.} & \textbf{R1$\uparrow$} & \textbf{mAP$\uparrow$} \\
\midrule
\multicolumn{5}{l}{\gray  \emph{\textbf{Score-Level Fusion}}} \\
\hdashline
F\,+\,G (Heavy)            & 693.6 & \faceE\ \gaitE & 91.4 & 85.7 \\
F\,+\,G (Light)             &  11.7 & \faceA\ \gaitA & 94.4 & 91.7 \\
F\,+\,B (Heavy)             &  36.8 & \faceE\ \bodyE & 93.6 & 90.0 \\
F\,+\,B (Light)             &  12.8 & \faceA\ \bodyA & 92.2 & 86.6 \\
G\,+\,B (Heavy)             & 681.8 & \gaitE\ \bodyE & 84.2 & 78.7 \\
G\,+\,B (Light)              &  14.1 & \gaitA\ \bodyA & 89.2 & 86.3 \\
F\,+\,B\,+\,G (Heavy)       & 706.1 & \faceE\ \bodyE\ \gaitE & 91.7 & 88.9 \\
F\,+\,B\,+\,G (Light)      &  19.4 & \faceA\ \bodyA\ \gaitA & 94.5 & 93.5 \\
\midrule
Asym-AO1~\cite{herbadji2020combining}  & 706.1 & \faceE\ \bodyA\ \gaitE & 92.3 & 90.0 \\
BSSF~\cite{teng2022optimized}         & 706.1 & \faceE\ \bodyA\ \gaitE & 91.8 & 91.1 \\
SapiensID*~\cite{kim2025sapiensid}    & -- & -- & 92.6 & 77.8 \\
Farsight~\cite{liu2024farsight}       & 706.1 & \faceE\ \bodyA\ \gaitE & 92.0 & 91.2 \\
QME~\cite{zhu2025quality}   & 706.1 & \faceE\ \bodyA\ \gaitE & 94.1 & 90.8 \\
\textbf{IDSelect} (Static)                      &  38.3 & \faceE\ \gaitA\ \bodyA & 94.6 & 93.8 \\
\textbf{IDSelect} (Adaptive)                      & \textbf{53.7} & Ada. & \textbf{95.9} & \textbf{94.6} \\
\bottomrule
\end{tabular}
}
\end{minipage}
\vspace{-3mm}
\end{table}

\Paragraph{Baselines.}
We compare against comprehensive baseline categories: (1) Single-modality approaches using individual face, gait, and body models, including BigGait~\cite{ye2024biggait}, which we use with weights trained on the CCPG long-range video dataset~\cite{Li_2023_CVPR}; (2) Advanced fixed fusion techniques such as Asym-AOI~\cite{herbadji2020combining}, BSSF~\cite{teng2022optimized}, and Farsight~\cite{liu2024farsight}; (3) Universal multi-modal models, including SapiensID~\cite{kim2025sapiensid}, a unified face–body model pre-trained on the WebBody4M dataset~\cite{kim2025sapiensid}; and (4) Quality-guided methods, including QME~\cite{zhu2025quality} (ICCV'25).

\Paragraph{Evaluation Metrics.} Following standard person re-identification protocols, we report Rank-1 accuracy and mean Average Precision (mAP) on fixed test sets. For computational efficiency analysis, we measure GFLOPs based on the model architecture and input size. 

\subsection{Results on CCVID Dataset}
Tab.~\ref{tab:ccvid1} presents results on CCVID \emph{Configuration 1}, where IDSelect demonstrates exceptional efficiency gains while improving accuracy. IDSelect (Adaptive) achieves 95.9\% Rank-1 and 94.6\% mAP while consuming only 53.7 GFLOPs, a 92.4\% reduction compared to QME's 706.1 GFLOPs with $+$1.8\% higher accuracy. This dramatic computational saving demonstrates that adaptive model selection can discover superior model combinations that heavyweight exhaustive fusion approaches fail to find.
Note that feature-level averaging across modalities is inapplicable in our setting, as different models produce embeddings in incompatible spaces and dimensions; score-level mean fusion (the Light/Heavy rows in Tab.~\ref{tab:ccvid1}) is therefore the appropriate simple baseline. IDSelect (Adaptive) outperforms even the strongest such baseline, F\,+\,B\,+\,G (Light), which achieves 94.5\% Rank-1 at 19.4 GFLOPs.

To further distinguish IDSelect from static selection approaches (\emph{e.g.}, NAS), we report IDSelect (Static) in Tab.~\ref{tab:ccvid1}, which deploys only the most frequently selected combination from the learned policy. This fixed combination achieves 94.6\%, 1.3\% below the adaptive variant's 95.9\%, confirming that per-input adaptive selection provides measurable gains that no fixed combination can match.

\begin{figure*}[t]
  \centering
   \includegraphics[width=0.98\linewidth]{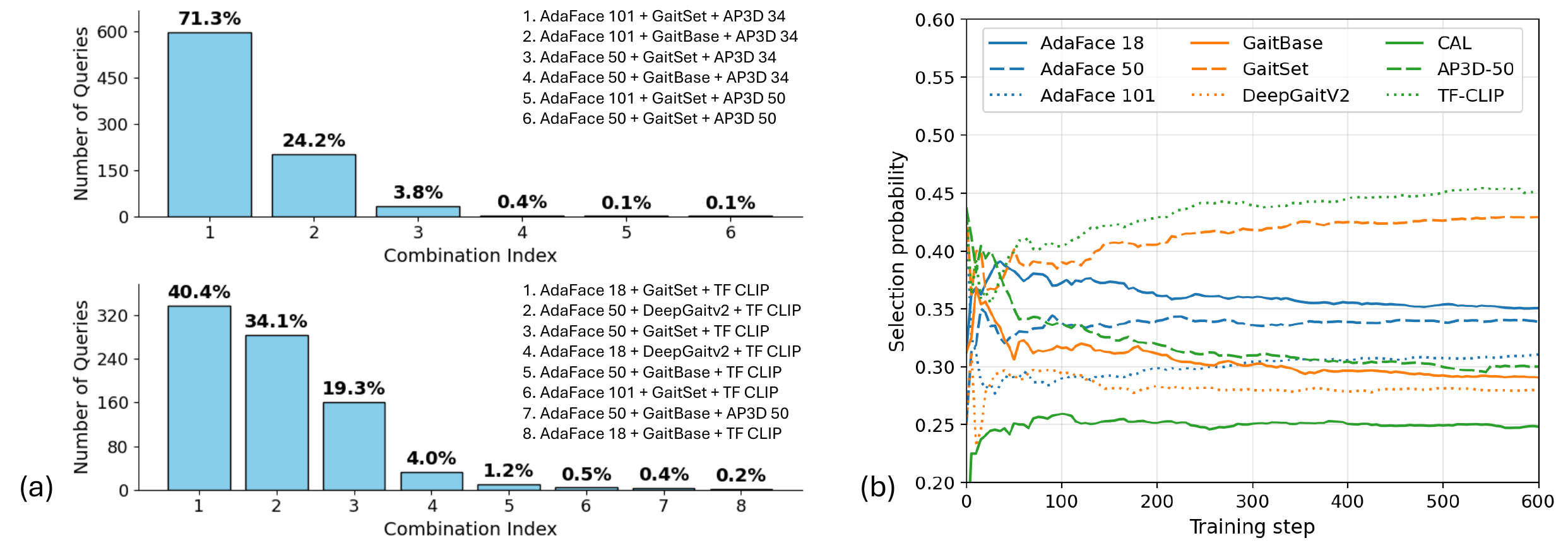}
  \caption{\textbf{(a)} Selection frequency distribution for CCVID configurations. \emph{Config. 1} shows concentrated patterns with IDSelect predominantly selecting \texttt{adaface\_101 + gaitset + ap3d\_34} (71.3\%) and \texttt{adaface\_101 + gaitbase + ap3d\_34} (24.2\%). \emph{Config. 2} exhibits more diverse selection patterns, demonstrating adaptability to different model pool characteristics. \textbf{(b)} Selection probability evolution during training (Config.~2). Body converges to TF-CLIP; Gait learns GaitSet/GaitBase preference; Face maintains input-adaptive diversity.} 
   \label{fig:ccvid_bar}
\end{figure*}

Tab.~\ref{tab:ccvid2} shows results on CCVID \emph{Config.~2} with a more constrained model pool. IDSelect achieves 97.4\% Rank-1 and 92.7\% mAP at 54.8 GFLOPs, a 2.2$\times$ savings over exhaustive heavy-model fusion (122.6 GFLOPs). Notably, \emph{Config.~2} achieves higher accuracy than \emph{Config.~1} (97.4\% vs 95.9\% Rank-1) despite tighter computational constraints, demonstrating IDSelect's adaptability to different model pool characteristics. This gap arises because \emph{Config.~2}'s TF-CLIP (ViT-based) produces sharper similarity distributions that boost top-1 discrimination but reduce ranking smoothness across the full gallery. The more diverse selection patterns (Fig.~\ref{fig:ccvid_bar} \textcolor{red}{(a)}) further optimize for challenging cases, benefiting Rank-1 while adding variance to mAP.

\begin{table}[t]
\centering

\begin{minipage}[t]{0.49\textwidth}
\centering
\captionof{table}{Performance comparison on CCVID \emph{Configuration 2}. IDSelect demonstrates adaptability across model pool configurations, maintaining strong performance while achieving significant computational savings compared to exhaustive fusion. \emph{w/o RL} uses the same selector and model pool with supervised training.}
\label{tab:ccvid2}

\resizebox{0.98\linewidth}{!}{%
\begin{tabular}{l c c c c c}
\toprule
\textbf{Model/Method} & \textbf{Arch.} & \textbf{GFLOPs$\downarrow$} & \textbf{Comb.} & \textbf{R1$\uparrow$} & \textbf{mAP$\uparrow$} \\
\midrule
\multicolumn{6}{l}{\gray  \emph{\textbf{Face Models*}}} \\ 
\hdashline
AdaFace18~\cite{kim2022adaface}      & R18     &  5.2  & \faceA & 90.5 & 80.4\\
AdaFace50~\cite{kim2022adaface}      & R50    &  12.7  & \faceC & 91.7 & 82.9\\
AdaFace101~\cite{kim2022adaface}     & R101  &  24.3   & \faceE & 94.0 & 87.9 \\ \hline
\multicolumn{6}{l}{\gray  \emph{\textbf{Body Models}}} \\
\hdashline
AP3D 50~\cite{gu2020appearance}         & R50      &  8.5   & \bodyA & 80.9 & 79.2 \\ 
CAL~\cite{CAL}         & R50    &   12.5  & \bodyC & 81.4 & 74.7 \\
TF CLIP~\cite{yu2024tf}     &  ViT-B16     &  14.8   & \bodyE & 90.8 & 90.4 \\ \hline
\multicolumn{6}{l}{\gray  \emph{\textbf{Gait Models}}} \\
\hdashline
GaitSet~\cite{chao2021gaitset}          &  CNN-6     &  6.5   & \gaitA & 79.3 & 73.8 \\ 
GaitBase~\cite{fan2023opengait}         & R9      &  71.0   & \gaitC & 64.0 & 61.4 \\ 
DeepGaitv2~\cite{fan2023exploring}  &  P3D-R*      & 83.5   & \gaitE & 68.5 & 59.9 \\
\midrule
\textbf{IDSelect} (\emph{w/o RL}) &   --    & 54.8   &    Ada.       & 94.8 & 92.5 \\
\textbf{IDSelect} &   --    & 54.8   &     Ada.      & \textbf{97.4} & \textbf{92.7} \\
\bottomrule
\end{tabular}}
\end{minipage} 

\vspace{3mm}

\begin{minipage}[t]{0.49\textwidth}
\centering
\captionof{table}{Performance comparison on MEVID dataset with face-body fusion configurations. Results highlight the benefits of cross-modal complementarity while revealing diminishing returns from intra-modal redundancy among body models.}
\label{tab:mevid}

\resizebox{0.98\linewidth}{!}{%
\begin{tabular}{l c c c c c}
\toprule
\textbf{Model/Method} & \textbf{Arch.} & \textbf{GFLOPs$\downarrow$} & \textbf{Comb.} & \textbf{R1$\uparrow$} & \textbf{mAP$\uparrow$} \\
\midrule
\multicolumn{6}{l}{\gray  \emph{\textbf{Face Models*}}} \\ 
\hdashline
AdaFace18~\cite{kim2022adaface}      & R18     & 5.2   & \faceA & 18.4 & 7.7\\
AdaFace50~\cite{kim2022adaface}      & R50     & 12.7   & \faceC & 23.1 & 9.1\\
AdaFace101~\cite{kim2022adaface}     & R101    & 24.3    & \faceE & 25.0 & 8.1 \\ \hline
\multicolumn{6}{l}{\gray  \emph{\textbf{Body Models}}} \\
\hdashline
AP3D~\cite{gu2020appearance}     & R50    &  8.5   & \bodyA & 35.4 & 14.6 \\
CAL~\cite{CAL}         & R50  &  12.5   & \bodyB & 52.5 & 27.1 \\
Attn-CL~\cite{pathak2020video}     & R50    &  12.7   & \bodyC & 44.0 & 16.1 \\
PiT~\cite{zang2022multidirection}         & ViT-B16    &  22.6   & \bodyD & 34.2 & 13.6 \\ 
AGRL~\cite{wu2020adaptive}     & R50    &  48.2   & \bodyE & 51.9 & 25.5 \\
\midrule
\multicolumn{6}{l}{\gray  \emph{\textbf{1 face model + 1 body model}}} \\ 
\hdashline
GEFF~\cite{arkushin2024geff}        & --       & 36.8    & \faceE\ \bodyB      & 32.9 & 18.8 \\
QME~\cite{zhu2025quality}        & --       & 36.8    &  \faceE\ \bodyB    & 33.5 & 19.9 \\
\textbf{IDSelect}       & --       &\textbf{28.7}     &  Ada.    &\textbf{50.0}    &\textbf{21.5}     \\
\midrule
\multicolumn{6}{l}{\gray  \emph{\textbf{1 face model + 2 body models}}} \\ 
\hdashline
Asym-AO1~\cite{herbadji2020combining}      &   --      &  85.0    & \faceE\ \bodyB\ \bodyE     & 52.5 & 22.9 \\
BSSF~\cite{teng2022optimized}      &    --     &  85.0    & \faceE\ \bodyB\ \bodyE     & 53.5 & 27.4 \\
Farsight~\cite{liu2024farsight}        & --       & 85.0    &   \faceE\ \bodyB\ \bodyE    & 53.8 & 25.4 \\
QME~\cite{zhu2025quality}  &    --   & 85.0   &    \faceE\ \bodyB\ \bodyE   & 55.7  & \textbf{28.2} \\
\textbf{IDSelect} &   --    & \textbf{49.9}   & Ada.     & \textbf{56.0} & 24.4 \\
\bottomrule
\end{tabular}}
\end{minipage}
\vspace{-3mm}
\end{table}

\subsection{Results on MEVID Dataset}

Tab.~\ref{tab:mevid} presents results on the challenging MEVID dataset with face-body fusion configurations. Following QME's experimental setup, we evaluate configurations with varying model pool sizes to demonstrate IDSelect's effectiveness under extreme viewing conditions. 
For the \textbf{1 face + 1 body model} configuration, IDSelect achieves 50.0\% Rank-1 and 21.5\% mAP at 28.7 GFLOPs. Notably, this is lower than the best single body model (52.5\% Rank-1), indicating that under MEVID's extreme viewing conditions, the degraded face modality (25.0\% Rank-1) actually reduces overall performance when fused with body features. In the \textbf{1 face + 2 body models} configuration, IDSelect reaches 56.0\% Rank-1 and 24.4\% mAP at 49.9 GFLOPs, outperforming QME's 55.7\% while achieving 1.7× computational savings.
The results reveal that under extreme surveillance conditions, face information can be too degraded to provide complementary benefits. However, the modest improvement when expanding from 1 to 2 body models (56.0\% vs 50.0\% Rank-1) suggests significant overlap among the body models, making it challenging to identify truly complementary combinations within the same modality. This finding validates IDSelect's design principle that optimal model selection benefits most from cross-modal diversity rather than intra-modal redundancy. Future work should explore face + gait + body fusion on datasets where all three modalities are reliable, as combining fundamentally different biometric cues offers the greatest complementarity potential.

\subsection{Analysis and Discussion}

\Paragraph{Modality Complementarity and Selection Patterns.}
To understand how IDSelect discovers complementary model combinations, we analyze performance across modality combinations and examine the learned selection patterns. Tab. 6 in supp reveals that cross-modal fusion consistently outperforms individual modalities, with the most significant gains occurring when combining fundamentally different biometric cues.
%
%
Fig.~\ref{fig:ccvid_bar} \textcolor{red}{(a)} reveals the learned selection patterns. In \emph{Config. 1}, IDSelect exhibits concentrated preferences, selecting \texttt{adaface\_101 + gaitset + ap3d\_34} in 71.3\% of cases, indicating that the agent prioritizes the most accurate face model while adaptively choosing gait models based on input characteristics. \emph{Config. 2} shows more diverse selections, demonstrating the framework's adaptability to model-pool characteristics.

\Paragraph{Limitations.}
IDSelect requires curating diverse model pools, incurring upfront costs not captured in inference metrics. However, this one-time investment becomes more practical as the community releases abundant pre-trained models (\emph{e.g.}, AdaFace, BigGait, SapiensID) that can be reused, turning pool construction into curation rather than training. While the framework must keep multiple models in memory, this overhead is acceptable in most scenarios where computational savings outweigh storage, and edge devices now support larger model capacities. Additionally, our experiments with small pools (3--5 models per modality) open opportunities to study scalability to larger ecosystems and domain adaptation of selection policies across deployments. Future work includes hierarchical selection strategies and backup mechanisms for cases where individual modalities fail under severe degradation.

\section{Conclusion}

We introduced \textbf{IDSelect}, an RL-based cost-aware agent that transforms video-based person recognition from static heavyweight fusion into adaptive resource allocation, achieving up to 92.4\% computational reduction while improving accuracy. The key insight is that learned selection policies uncover complementary cross-modal synergies that fixed approaches cannot, especially the finding that diversity across face, gait, and body yields greater gains than intra-modal redundancy. By balancing recognition performance and computational cost through input-conditioned model selection, IDSelect enables the practical deployment of multi-modal video biometrics in resource-constrained settings, shifting the paradigm from ``use everything available'' to ``select only what is needed.''


\Paragraph{Acknowledgments} This work was supported in part by the National Science Foundation under Grant FMitF-2319242.
\clearpage
\section*{Supplementary Material}


This supplementary material provides implementation details,
additional analyses of modality complementarity, policy behavior,
efficiency and memory, RL variants, hyperparameter sensitivity,
and qualitative examples.

\section{Implementation Details}

\paragraph{Selection Agent Architecture.}
We implement the selection agent with a lightweight CNN encoder featuring five convolutional layers
(3$\rightarrow$64$\rightarrow$128$\rightarrow$256$\rightarrow$512 channels),
followed by adaptive pooling and projection to 512-dimensional features.
Modality-specific agent heads are implemented as 3-layer MLPs
(512$\rightarrow$256$\rightarrow$128 dimensions) that output selection
logits for each modality's available models.

\paragraph{Optimization Settings.}
Reinforcement learning uses the Adam optimizer with learning rate
$5 \times 10^{-4}$, batch size $8$, and weight decay
$1 \times 10^{-4}$. We apply gradient clipping with max norm $1.0$,
entropy coefficient $\beta = 0.01$, critic weight $\alpha = 0.5$,
and target normalized cost $C_{\text{target}} = 0.45$.
The Lagrange multiplier is initialized at $0.1$ and updated with
step size $5 \times 10^{-3}$.

\paragraph{Gallery Processing.}
At inference, gallery embeddings are precomputed for each candidate model. The model selected for the probe determines the corresponding gallery embeddings used for matching, avoiding per-query gallery-side recomputation. During training, the selected model(s) are applied consistently to both query and gallery samples for feature extraction.

\section{Additional Analysis}

In Tab.~\ref{tab:ccvid_model1_ablation}, Face + Gait achieves 95.5\% Rank-1, demonstrating strong complementarity between facial features and temporal gait patterns. Face + Body fusion achieves 94.2\% Rank-1, showing efficient performance when gait analysis is unavailable. Gait + Body achieves 87.5\% Rank-1, indicating that body shape and gait dynamics provide reasonable recognition even without facial information. The three-modality approach achieves optimal performance (95.9\% Rank-1, 94.6\% mAP), where the significant mAP improvement justifies the additional computational cost.

\paragraph{RL Policy Comparison.}
We chose Actor-Critic for its simplicity and stable integration with the Lagrangian budget controller. To validate this choice, we additionally tested PPO: it achieves comparable results (96.1\% Rank-1 vs.\ 95.9\% on Config 1) but adds clipping complexity unnecessary for our small action space (27 combinations). A3C requires asynchronous infrastructure; GRPO targets LLM alignment with group-relative rewards, a fundamentally different setting. Our framework is modular and the base RL algorithm can be swapped without architectural changes.

\vspace{1mm}
\paragraph{Efficiency Trade-offs.}
IDSelect consistently discovers optimal efficiency regions unavailable to fixed strategies. As shown in Tabs.~\ref{tab:ccvid1} and~\ref{tab:ccvid2}, (Light) combinations achieve reasonable efficiency but lower accuracy (F\,+\,B\,+\,G (Light): 94.5\% Rank-1 at 19.4 GFLOPs), while (Heavy) combinations yield varying accuracy at high computational cost (F\,+\,B\,+\,G (Heavy): 91.7\% Rank-1 at 706.1 GFLOPs). IDSelect finds superior trade-offs with 95.9--97.4\% Rank-1 at moderate cost (53.7--54.8 GFLOPs), achieving a 13.2$\times$ reduction on CCVID \emph{Config.~1} while improving accuracy from 94.1\% to 95.9\%; a 2.2$\times$ reduction on \emph{Config.~2}; and 1.7$\times$ savings on MEVID while maintaining comparable Rank-1 accuracy.

\paragraph{Memory Analysis.}
In full dynamic mode, the 9-model pool (3 per modality) requires $\sim$5.2~GB total VRAM, loaded once at initialization with no runtime swapping. The selection agent adds $<$24~MB overhead, modest relative to the 92.4\% FLOP reduction, which provides substantial latency savings. For reference, the ablation IDSelect (Static) in Tab.~\ref{tab:ccvid1} shows that a single fixed combination recovers most gains (94.6\% vs.\ 95.9\% Rank-1) at standard 3-model memory, quantifying the 1.3\% cost of forgoing adaptivity.

\paragraph{Hardware-Level Efficiency.}

\begin{table}[t]
\centering
\small
\renewcommand{\arraystretch}{1.0}
\setlength{\tabcolsep}{5pt}
\caption{Hardware-level efficiency on an NVIDIA A100.}
\label{tab:hardware_efficiency}
\resizebox{0.95\linewidth}{!}{
\begin{tabular}{lcccc}
\toprule
\textbf{Method} &
\textbf{GFLOPs$\downarrow$} &
\textbf{Lat. (ms)$\downarrow$} &
\textbf{Thr. (q/s)$\uparrow$} &
\textbf{VRAM (GB)$\downarrow$} \\
\midrule
QME & 706.1 & 71.3 & 14.0 & 3.6 \\
F+B+G (Light) & 19.4 & 10.5 & 94.9 & 1.2 \\
IDSelect & 53.7 & 26.1 & 38.4 & 5.2 \\
\bottomrule
\end{tabular}
}
\end{table}

We measure end-to-end query latency, throughput, and peak GPU memory on an NVIDIA A100. Compared with QME, IDSelect achieves $2.7\times$ lower latency and $2.7\times$ higher throughput while improving Rank-1 by 1.8 points. The lightweight fixed combination remains faster, reflecting the trade-off between minimum computation and the additional accuracy obtained through adaptive selection.

\paragraph{Comparison with Non-RL Routing.}
To isolate the effect of RL optimization from the selector architecture, we compare IDSelect with heuristic and supervised routing under the same CCVID \emph{Configuration 2} model pool and a comparable computational budget.The heuristic router independently estimates a quality score for each
modality using face confidence, body completeness, and gait consistency.
For each modality, the quality scores are divided into three quantile
bins (low, medium, and high). Low-quality samples are routed to the
highest-capacity model, medium-quality samples to the intermediate
model, and high-quality samples to the lightweight model. The resulting
face, body, and gait selections are then combined using the same
multi-modal fusion protocol as IDSelect.
\begin{table}[t]
\centering
\small
\renewcommand{\arraystretch}{1.0}
\setlength{\tabcolsep}{5pt}
\caption{Comparison with non-RL routing on CCVID
\emph{Configuration 2}.}
\label{tab:nonrl}
\resizebox{0.80\linewidth}{!}{
\begin{tabular}{lccc}
\toprule
\textbf{Selector} &
\textbf{GFLOPs$\downarrow$} &
\textbf{Rank-1$\uparrow$} &
\textbf{mAP$\uparrow$} \\
\midrule
Heuristic & 52.6 & 92.3 & 87.4 \\
Supervised & 54.8 & 94.8 & 92.5 \\
IDSelect (RL) & 54.8 & \textbf{97.4} & \textbf{92.7} \\
\bottomrule
\end{tabular}
}
\end{table}
The supervised variant uses the same selector architecture and model
pool as IDSelect but replaces the RL objective with supervised training.
Target actions are constructed by evaluating all candidate combinations
for each training sample, and argmax selection is used at inference. At the same 54.8 GFLOPs, RL improves Rank-1 from 94.8\% to 97.4\%, while the mAP improvement is smaller (92.5\% to 92.7\%). The heuristic strategy performs substantially worse.

\paragraph{Additional Evaluation on SUSTech1K.}

\begin{table}[t]
\centering
\small
\renewcommand{\arraystretch}{1.0}
\setlength{\tabcolsep}{5pt}
\caption{Additional evaluation on SUSTech1K.}
\label{tab:sustech}
\resizebox{0.50\linewidth}{!}{
\begin{tabular}{lc}
\toprule
\textbf{Setting} & \textbf{Rank-1$\uparrow$} \\
\midrule
Gait only & 81.9 \\
Body only & 85.7 \\
IDSelect (Body + Gait) & \textbf{96.3} \\
\bottomrule
\end{tabular}
}
\end{table}

To examine generalization beyond CCVID and MEVID, we evaluate IDSelect on SUSTech1K using Body and Gait modalities. The Body+Gait configuration achieves 96.3\% Rank-1, compared with 81.9\% for Gait-only and 85.7\% for Body-only, providing additional evidence of cross-modal complementarity beyond the primary CCVID and MEVID benchmarks.

\paragraph{Selection Policy Evolution.}
Fig.~\ref{fig:ccvid_bar} \textcolor{red}{(b)} visualizes how selection probabilities evolve during training on \emph{Config.~2}. Body selection quickly converges to TF-CLIP (the strongest body model), while Gait develops a clear preference for GaitSet/GaitBase. Face selection remains diverse---the policy selects different face models for different inputs (\emph{e.g.}, AdaFace18 $\sim$40\%, AdaFace50 $\sim$53\% in Fig.~\ref{fig:ccvid_bar}), so the marginal probability stays balanced while per-input selection is decisive. This confirms input-adaptive rather than random behavior.

\section{Ablation Studies}
\label{sec:s_ablation}

\subsection{Hyperparameter Analysis}
We analyze sensitivity to key hyperparameters in the RL training process.

\begin{table}[h]
    \centering
    \renewcommand{\arraystretch}{1.05}
    \caption{Impact of entropy coefficient $\beta$ (CCVID Config 2).}
    \label{tab:s_ablation_beta}
    \begin{tabular}{lcc}
        \toprule
        $\beta$ & Rank-1 & mAP \\
        \midrule
        0.001 & 97.0 & 91.8 \\
        0.1 & 96.5 & 90.5 \\
        \midrule
        0.01 (ours) & \textbf{97.4} & \textbf{92.7} \\
        \bottomrule
    \end{tabular}
\end{table}

\begin{table}[h]
    \centering
    \renewcommand{\arraystretch}{1.05}
    \caption{Impact of target cost budget $C_{\text{target}}$ (CCVID Config 2).}
    \label{tab:s_ablation_budget}
    \begin{tabular}{lcc}
        \toprule
        $C_{\text{target}}$ & Rank-1 & mAP \\
        \midrule
        0.35 & 95.8 & 89.5 \\
        0.50 & 97.0 & 92.0 \\
        \midrule
        0.45 (ours) & \textbf{97.4} & \textbf{92.7} \\
        \bottomrule
    \end{tabular}
\end{table}

\subsection{Hyperparameter Sensitivity Analysis}

The ablation results in Tab.~\ref{tab:s_ablation_beta} and Tab.~\ref{tab:s_ablation_budget} reveal important insights about the robustness of IDSelect.

\paragraph{Entropy coefficient $\beta$.} With $\beta=0.001$, the policy achieves 97.0\% Rank-1 but a lower mAP (91.8\%), suggesting premature convergence to suboptimal model combinations. At $\beta=0.1$, excessive exploration leads to degraded performance (96.5\% Rank-1, 90.5\% mAP). Our choice of $\beta=0.01$ achieves the best balance (97.4\% Rank-1 and 92.7\% mAP).

\paragraph{Target cost budget $C_{\text{target}}$.} At $C_{\text{target}}=0.35$, the aggressive budget constraint forces lightweight model selection, achieving only 95.8\% Rank-1 and 89.5\% mAP. Relaxing the budget to $C_{\text{target}}=0.50$ enables heavier models (97.0\% Rank-1, 92.0\% mAP); however, the performance gain plateaus as the agent may select computationally expensive models that are not necessarily the most complementary, exhibiting diminishing returns. Our setting of $C_{\text{target}}=0.45$ discovers the optimal operating point (97.4\% Rank-1 and 92.7\% mAP), demonstrating that peak performance does not require maximizing computational resources but rather finding the right balance.


\begin{figure}[t]
    \centering
    \includegraphics[width=0.7\linewidth]{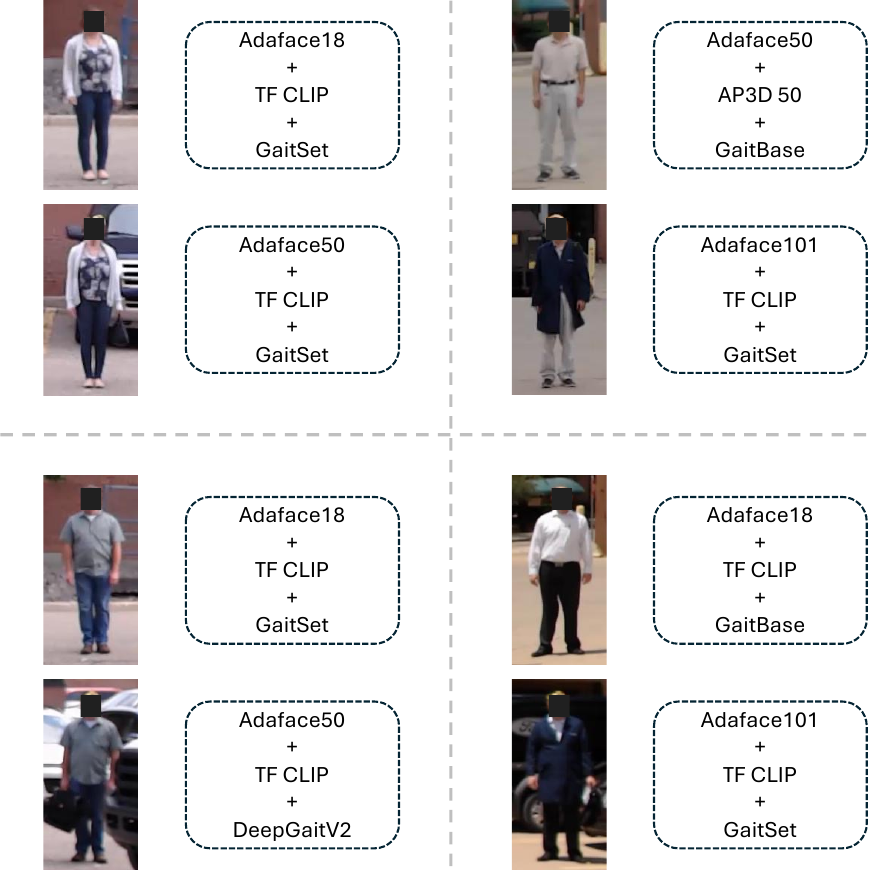}
    \caption{Additional qualitative results. The agent selects light models for clean inputs and escalates to stronger face/body/gait experts under challenging conditions.}
    \label{fig:s_qual_examples}
\end{figure}

\begin{table}[t]
\centering
\setlength{\tabcolsep}{8pt}
\renewcommand{\arraystretch}{1.0}
\caption{Modality combination analysis on CCVID \emph{Configuration 1}. Results demonstrate the effectiveness of cross-modal complementarity, with the full three-modality approach achieving optimal performance while two-modality combinations provide competitive accuracy at reduced computational cost.}
\resizebox{0.75\linewidth}{!}{
\begin{tabular}{l c c c} 
\toprule
\textbf{Method} & \textbf{GFLOPs$\downarrow$} & \textbf{Rank1$\uparrow$} & \textbf{mAP$\uparrow$} \\
\midrule
\midrule
Face\,+\,Gait   & 46.1 & 95.5 & 91.2 \\
Face\,+\,Body   & 31.4 & 94.2 & 90.3 \\
Gait\,+\,Body   & 29.9 & 87.5 & 85.1 \\
Face\,+\,Gait\,+\,Body   & 53.7 & 95.9 & 94.6 \\
\bottomrule
\end{tabular}
}
\label{tab:ccvid_model1_ablation}
\end{table}


{
    \small
    \bibliographystyle{ieeenat_fullname}
    \bibliography{main}

@inproceedings{arkushin2024geff,
  author    = {Arkushin, Daniel and Cohen, Bar and Peleg, Shmuel and Fried, Ohad},
  title     = {{GEFF}: Improving any clothes-changing person {ReID} model using gallery enrichment with face features},
  booktitle = {WACVW},
  year      = {2024},
}

@article{aung2022multimodal,
  author    = {Aung, Hsu Mon Lei and Pluempitiwiriyawej, Charnchai and Hamamoto, Kazuhiko and Wangsiripitak, Somkiat},
  title     = {Multimodal biometrics recognition using a deep convolutional neural network with transfer learning in surveillance videos},
  journal   = {Computation},
  volume    = {10},
  number    = {7},
  pages     = {127},
  year      = {2022},
}

@inproceedings{bolukbasi2017adaptive,
  author    = {Bolukbasi, Tolga and Wang, Joseph and Dekel, Ofer and Saligrama, Venkatesh},
  title     = {Adaptive neural networks for efficient inference},
  booktitle = {ICML},
  year      = {2017},
}

@article{chao2021gaitset,
  author    = {Chao, Hanqing and Wang, Kun and He, Yiwei and Zhang, Junping and Feng, Jianfeng},
  title     = {{GaitSet}: Cross-view gait recognition through utilizing gait as a deep set},
  journal   = {IEEE TPAMI},
  volume    = {44},
  number    = {7},
  pages     = {3467--3478},
  year      = {2022},
}

@inproceedings{davila2023mevid,
  author    = {Davila, Daniel and Du, Dawei and Lewis, Bryon and Funk, Christopher and Van Pelt, Joseph and Collins, Roderic and Corona, Kellie and Brown, Matt and McCloskey, Scott and Hoogs, Anthony and Clipp, Brian},
  title     = {{MEVID}: Multi-view extended videos with identities for video person re-identification},
  booktitle = {WACV},
  year      = {2023},
}

@inproceedings{deng2019arcface,
  author    = {Deng, Jiankang and Guo, Jia and Xue, Niannan and Zafeiriou, Stefanos},
  title     = {{ArcFace}: Additive angular margin loss for deep face recognition},
  booktitle = {CVPR},
  year      = {2019},
}

@article{fan2023exploring,
  author    = {Fan, Chao and Hou, Saihui and Huang, Yongzhen and Yu, Shiqi},
  title     = {Exploring deep models for practical gait recognition},
  journal   = {arXiv preprint arXiv:2303.03301},
  year      = {2023},
}

@inproceedings{fan2023opengait,
  author    = {Fan, Chao and Liang, Junhao and Shen, Chuanfu and Hou, Saihui and Huang, Yongzhen and Yu, Shiqi},
  title     = {{OpenGait}: Revisiting gait recognition towards better practicality},
  booktitle = {CVPR},
  year      = {2023},
}

@article{george2024edgeface,
  author    = {George, Anjith and Ecabert, Christophe and Otroshi Shahreza, Hatef and Kotwal, Ketan and Marcel, S{\'e}bastien},
  title     = {{EdgeFace}: Efficient face recognition model for edge devices},
  journal   = {IEEE T-BIOM},
  volume    = {6},
  number    = {2},
  pages     = {158--168},
  year      = {2024},
}

@incollection{gong2011person,
  author    = {Gong, Shaogang and Xiang, Tao},
  title     = {Person re-identification},
  booktitle = {Visual Analysis of Behaviour: From Pixels to Semantics},
  pages     = {301--313},
  publisher = {Springer},
  address   = {London},
  year      = {2011},
}

@inproceedings{gu2020appearance,
  author    = {Gu, Xinqian and Chang, Hong and Ma, Bingpeng and Zhang, Hongkai and Chen, Xilin},
  title     = {Appearance-preserving {3D} convolution for video-based person re-identification},
  booktitle = {ECCV},
  year      = {2020},
}

@inproceedings{CAL,
  author    = {Gu, Xinqian and Chang, Hong and Ma, Bingpeng and Bai, Shutao and Shan, Shiguang and Chen, Xilin},
  title     = {Clothes-changing person re-identification with {RGB} modality only},
  booktitle = {CVPR},
  year      = {2022},
}

@article{he2024efficient,
  author    = {He, Yifei and Cheng, Runxiang and Balasubramaniam, Gargi and Tsai, Yao-Hung Hubert and Zhao, Han},
  title     = {Efficient modality selection in multimodal learning},
  journal   = {JMLR},
  volume    = {25},
  number    = {47},
  pages     = {1--39},
  year      = {2024},
}

@article{herbadji2020combining,
  author    = {Herbadji, Abderrahmane and Akhtar, Zahid and Siddique, Kamran and Guermat, Noubeil and Ziet, Lahcene and Cheniti, Mohamed and Muhammad, Khan},
  title     = {Combining multiple biometric traits using asymmetric aggregation operators for improved person recognition},
  journal   = {Symmetry},
  volume    = {12},
  number    = {3},
  pages     = {444},
  year      = {2020},
}

@article{huang2024whole,
  author    = {Huang, Siyuan and Kathirvel, Ram Prabhakar and Guo, Yuxiang and Lau, Chun Pong and Chellappa, Rama},
  title     = {Whole-body detection, identification and recognition at altitude and range},
  journal   = {IEEE T-BIOM},
  volume    = {7},
  number    = {3},
  pages     = {331--343},
  year      = {2025},
}

@article{jain2005score,
  author    = {Jain, Anil and Nandakumar, Karthik and Ross, Arun},
  title     = {Score normalization in multimodal biometric systems},
  journal   = {Pattern Recognition},
  volume    = {38},
  number    = {12},
  pages     = {2270--2285},
  year      = {2005},
}

@book{jain2024speaker,
  author    = {Jain, Anil K. and Ross, Arun A. and Nandakumar, Karthik and Swearingen, Thomas},
  title     = {Introduction to Biometrics},
  edition   = {Second},
  publisher = {Springer},
  address   = {Cham},
  year      = {2025},
}

@article{jawade2024conan,
  author    = {Jawade, Bhavin and Mohan, Deen Dayal and Shetty, Prajwal and Fedorishin, Dennis and Setlur, Srirangaraj and Govindaraju, Venu},
  title     = {{CoNAN}: Conditional neural aggregation network for unconstrained long range biometric feature fusion},
  journal   = {IEEE T-BIOM},
  volume    = {6},
  number    = {4},
  pages     = {602--612},
  year      = {2024},
}

@inproceedings{kim2022adaface,
  author    = {Kim, Minchul and Jain, Anil K. and Liu, Xiaoming},
  title     = {{AdaFace}: Quality adaptive margin for face recognition},
  booktitle = {CVPR},
  year      = {2022},
}

@inproceedings{kim2025sapiensid,
  author    = {Kim, Minchul and Ye, Dingqiang and Su, Yiyang and Liu, Feng and Liu, Xiaoming},
  title     = {{SapiensID}: Foundation for human recognition},
  booktitle = {CVPR},
  year      = {2025},
}

@inproceedings{Li_2023_CVPR,
  author    = {Li, Weijia and Hou, Saihui and Zhang, Chunjie and Cao, Chunshui and Liu, Xu and Huang, Yongzhen and Zhao, Yao},
  title     = {An in-depth exploration of person re-identification and gait recognition in cloth-changing conditions},
  booktitle = {CVPR},
  year      = {2023},
}

@inproceedings{li2019airface,
  author    = {Li, Xianyang and Wang, Feng and Hu, Qinghao and Leng, Cong},
  title     = {{AirFace}: Lightweight and efficient model for face recognition},
  booktitle = {ICCVW},
  year      = {2019},
}

@inproceedings{liu2024farsight,
  author    = {Liu, Feng and Ashbaugh, Ryan and Chimitt, Nicholas and Hassan, Najmul and Hassani, Ali and Jaiswal, Ajay and Kim, Minchul and Mao, Zhiyuan and Perry, Christopher and Ren, Zhiyuan and Su, Yiyang and Varghaei, Pegah and Wang, Kai and Zhang, Xingguang and Chan, Stanley and Ross, Arun and Shi, Humphrey and Wang, Zhangyang and Jain, Anil and Liu, Xiaoming},
  title     = {{FarSight}: A physics-driven whole-body biometric system at large distance and altitude},
  booktitle = {WACV},
  year      = {2024},
}

@article{liu2025person,
  author    = {Liu, Feng and Chimitt, Nicholas and Guo, Lanqing and Jain, Jitesh and Kane, Aditya and Kim, Minchul and Robbins, Wes and Su, Yiyang and Ye, Dingqiang and Zhang, Xingguang and Zhu, Jie and Satyakam, Siddharth and Perry, Christopher and Chan, Stanley H. and Ross, Arun and Shi, Humphrey and Wang, Zhangyang and Jain, Anil and Liu, Xiaoming},
  title     = {Person recognition at altitude and range: Fusion of face, body shape and gait},
  journal   = {arXiv preprint arXiv:2505.04616},
  year      = {2025},
}

@inproceedings{meng2022adavit,
  author    = {Meng, Lingchen and Li, Hengduo and Chen, Bor-Chun and Lan, Shiyi and Wu, Zuxuan and Jiang, Yu-Gang and Lim, Ser-Nam},
  title     = {{AdaViT}: Adaptive vision transformers for efficient image recognition},
  booktitle = {CVPR},
  year      = {2022},
}

@article{nandakumar2008likelihood,
  author    = {Nandakumar, Karthik and Chen, Yi and Dass, Sarat C. and Jain, Anil K.},
  title     = {Likelihood ratio-based biometric score fusion},
  journal   = {IEEE TPAMI},
  volume    = {30},
  number    = {2},
  pages     = {342--347},
  year      = {2008},
}

@inproceedings{panda2021adamml,
  author    = {Panda, Rameswar and Chen, Chun-Fu Richard and Fan, Quanfu and Sun, Ximeng and Saenko, Kate and Oliva, Aude and Feris, Rogerio},
  title     = {{AdaMML}: Adaptive multi-modal learning for efficient video recognition},
  booktitle = {ICCV},
  year      = {2021},
}

@inproceedings{pathak2020video,
  author    = {Pathak, Priyank and Eshratifar, Amir Erfan and Gormish, Michael},
  title     = {Video person {Re-ID}: Fantastic techniques and where to find them (student abstract)},
  booktitle = {AAAI},
  year      = {2020},
}

@inproceedings{rao2017attention,
  author    = {Rao, Yongming and Lu, Jiwen and Zhou, Jie},
  title     = {Attention-aware deep reinforcement learning for video face recognition},
  booktitle = {ICCV},
  year      = {2017},
}

@article{sun2025flexireid,
  author    = {Sun, Zhen and Tan, Lei and Shen, Yunhang and Cai, Chengmao and Sun, Xing and Dai, Pingyang and Cao, Liujuan and Ji, Rongrong},
  title     = {{FlexiReID}: Adaptive mixture of expert for multi-modal person re-identification},
  journal   = {arXiv preprint arXiv:2510.15595},
  year      = {2025},
}

@inproceedings{teerapittayanon2016branchynet,
  author    = {Teerapittayanon, Surat and McDanel, Bradley and Kung, Hsiang-Tsung},
  title     = {{BranchyNet}: Fast inference via early exiting from deep neural networks},
  booktitle = {ICPR},
  year      = {2016},
}

@article{teng2022optimized,
  author    = {Teng, Jackson Horlick and Ong, Thian Song and Connie, Tee and Sonai Muthu Anbananthen, Kalaiarasi and Min, Pa Pa},
  title     = {Optimized score level fusion for multi-instance finger vein recognition},
  journal   = {Algorithms},
  volume    = {15},
  number    = {5},
  pages     = {161},
  year      = {2022},
}

@article{wu2020adaptive,
  author    = {Wu, Yiming and Bourahla, Omar El Farouk and Li, Xi and Wu, Fei and Tian, Qi and Zhou, Xue},
  title     = {Adaptive graph representation learning for video person re-identification},
  journal   = {IEEE TIP},
  volume    = {29},
  pages     = {8821--8830},
  year      = {2020},
}

@inproceedings{yan2019vargfacenet,
  author    = {Yan, Mengjia and Zhao, Mengao and Xu, Zining and Zhang, Qian and Wang, Guoli and Su, Zhizhong},
  title     = {{VarGFaceNet}: An efficient variable group convolutional neural network for lightweight face recognition},
  booktitle = {ICCVW},
  year      = {2019},
}

@inproceedings{yang2020resolution,
  author    = {Yang, Le and Han, Yizeng and Chen, Xi and Song, Shiji and Dai, Jifeng and Huang, Gao},
  title     = {Resolution adaptive networks for efficient inference},
  booktitle = {CVPR},
  year      = {2020},
}

@inproceedings{ye2024biggait,
  author    = {Ye, Dingqiang and Fan, Chao and Ma, Jingzhe and Liu, Xiaoming and Yu, Shiqi},
  title     = {{BigGait}: Learning gait representation you want by large vision models},
  booktitle = {CVPR},
  year      = {2024},
}

@inproceedings{yin2022vit,
  author    = {Yin, Hongxu and Vahdat, Arash and Alvarez, Jose M. and Mallya, Arun and Kautz, Jan and Molchanov, Pavlo},
  title     = {{A-ViT}: Adaptive tokens for efficient vision transformer},
  booktitle = {CVPR},
  year      = {2022},
}

@inproceedings{yu2024tf,
  author    = {Yu, Chenyang and Liu, Xuehu and Wang, Yingquan and Zhang, Pingping and Lu, Huchuan},
  title     = {{TF-CLIP}: Learning text-free {CLIP} for video-based person re-identification},
  booktitle = {AAAI},
  year      = {2024},
}

@inproceedings{yun2024flex,
  author    = {Yun, Sukwon and Choi, Inyoung and Peng, Jie and Wu, Yangfan and Bao, Jingxuan and Zhang, Qiyiwen and Xin, Jiayi and Long, Qi and Chen, Tianlong},
  title     = {{Flex-MoE}: Modeling arbitrary modality combination via the flexible mixture-of-experts},
  booktitle = {NeurIPS},
  year      = {2024},
}

@article{zang2022multidirection,
  author    = {Zang, Xianghao and Li, Ge and Gao, Wei},
  title     = {Multidirection and multiscale pyramid in transformer for video-based pedestrian retrieval},
  journal   = {IEEE TII},
  volume    = {18},
  number    = {12},
  pages     = {8776--8785},
  year      = {2022},
}

@inproceedings{zhang2025router,
  author    = {Zhang, Haozhen and Feng, Tao and You, Jiaxuan},
  title     = {{Router-R1}: Teaching {LLMs} multi-round routing and aggregation via reinforcement learning},
  booktitle = {NeurIPS},
  year      = {2025},
}

@article{zheng2016person,
  author    = {Zheng, Liang and Yang, Yi and Hauptmann, Alexander G.},
  title     = {Person re-identification: Past, present and future},
  journal   = {arXiv preprint arXiv:1610.02984},
  year      = {2016},
}

@inproceedings{zhu2024sharc,
  author    = {Zhu, Haidong and Zheng, Wanrong and Zheng, Zhaoheng and Nevatia, Ram},
  title     = {{ShARc}: Shape and appearance recognition for person identification in-the-wild},
  booktitle = {WACV},
  year      = {2024},
}

@inproceedings{zhu2025quality,
  author    = {Zhu, Jie and Su, Yiyang and Kim, Minchul and Jain, Anil and Liu, Xiaoming},
  title     = {A quality-guided mixture of score-fusion experts framework for human recognition},
  booktitle = {ICCV},
  year      = {2025},
}

@inproceedings{su2025hamobe,
  author    = {Su, Yiyang and Shi, Yunping and Liu, Feng and Liu, Xiaoming},
  title     = {{HAMoBE}: Hierarchical and adaptive mixture of biometric experts for video-based person {ReID}},
  booktitle = {ICCV},
  year      = {2025},
}

@inproceedings{liu2023learning,
  author    = {Liu, Feng and Kim, Minchul and Gu, ZiAng and Jain, Anil and Liu, Xiaoming},
  title     = {Learning clothing and pose invariant {3D} shape representation for long-term person re-identification},
  booktitle = {ICCV},
  year      = {2023},
}

@inproceedings{liu2024distilling,
  author    = {Liu, Feng and Kim, Minchul and Ren, Zhiyuan and Liu, Xiaoming},
  title     = {Distilling {CLIP} with dual guidance for learning discriminative human body shape representation},
  booktitle = {CVPR},
  year      = {2024},
}

@article{zhang2020learning,
  author    = {Zhang, Ziyuan and Tran, Luan and Liu, Feng and Liu, Xiaoming},
  title     = {On learning disentangled representations for gait recognition},
  journal   = {IEEE TPAMI},
  volume    = {44},
  number    = {1},
  pages     = {345--360},
  year      = {2022},
}

@inproceedings{kim2024keypoint,
  author    = {Kim, Minchul and Su, Yiyang and Liu, Feng and Jain, Anil and Liu, Xiaoming},
  title     = {{KeyPoint} relative position encoding for face recognition},
  booktitle = {CVPR},
  year      = {2024},
}

@inproceedings{kim2022cluster,
  author    = {Kim, Minchul and Liu, Feng and Jain, Anil K. and Liu, Xiaoming},
  title     = {Cluster and aggregate: Face recognition with large probe set},
  booktitle = {NeurIPS},
  year      = {2022},
}

@inproceedings{su2026sapiensid2,
  author    = {Su, Yiyang and Zhu, Jie and Liu, Feng and Jain, Anil K. and Liu, Xiaoming},
  title     = {{SapiensID} 2.0: Aligning human recognition foundation models with human perception},
  booktitle = {NeurIPS},
  year      = {2026},
}

@article{nguyen2026aerialsurvey,
  author    = {Nguyen, Kien and Liu, Feng and Fookes, Clinton and Sridharan, Sridha and Liu, Xiaoming and Ross, Arun},
  title     = {Person recognition in aerial surveillance: A decade survey},
  journal   = {IEEE T-BIOM},
  volume    = {8},
  number    = {1},
  pages     = {3--19},
  year      = {2026},
}

@inproceedings{nguyen2025agvpreid,
  author    = {Nguyen, Huy and Nguyen, Kien and Pemasiri, Akila and Liu, Feng and Sridharan, Sridha and Fookes, Clinton},
  title     = {{AG-VPReID}: A challenging large-scale benchmark for aerial-ground video-based person re-identification},
  booktitle = {CVPR},
  year      = {2025},
}

@inproceedings{liu2022cfsm,
  author    = {Liu, Feng and Kim, Minchul and Jain, Anil and Liu, Xiaoming},
  title     = {Controllable and guided face synthesis for unconstrained face recognition},
  booktitle = {ECCV},
  year      = {2022},
}

@String(CVPR= {IEEE Conf. Comput. Vis. Pattern Recog.})

@String(ICCV= {Int. Conf. Comput. Vis.})

@String(ECCV= {Eur. Conf. Comput. Vis.})

@String(ICPR = {Int. Conf. Pattern Recog.})

@String(TIP  = {IEEE Trans. Image Process.})

@String(AAAI = {AAAI})

@String(CVPR  = {CVPR})

@String(ICCV  = {ICCV})

@String(ECCV  = {ECCV})

@String(ICPR  = {ICPR})

@String(TIP   = {IEEE TIP})
}

\end{document}